\documentclass[final]{anthology-ch-arxiv}

\usepackage{booktabs}
\usepackage{graphicx}
\usepackage{graphicx} % in preamble
\usepackage{makecell}
\usepackage{svg}
\usepackage{hyperref}
\usepackage{tablefootnote}
\usepackage{subcaption}
\usepackage{pdfpages}
\usepackage{float}

\usepackage{multirow}

\title{Llama-Embed-Nemotron-8B: A Universal Text Embedding Model for Multilingual and Cross-Lingual Tasks}

\begin{document}

\maketitle

\begin{center}

Yauhen~Babakhin, Radek~Osmulski, Ronay~Ak, Gabriel~Moreira, Mengyao~Xu, Benedikt~Schifferer, Bo~Liu, Even~Oldridge

\leavevmode\\

NVIDIA\footnote{Correspondence to Yauhen Babakhin (ybabakhin@nvidia.com)}
\end{center}

\begin{abstract}

We introduce llama-embed-nemotron-8b, an open-weights text embedding model\footnote{We released the model at https://huggingface.co/nvidia/llama-embed-nemotron-8b} that achieves state-of-the-art performance on the Multilingual Massive Text Embedding Benchmark (MMTEB) leaderboard as of October 21, 2025. While recent models show strong performance, their training data or methodologies are often not fully disclosed. We aim to address this by developing a fully open-source model, publicly releasing its weights and detailed ablation studies, and planning to share the curated training datasets. Our model demonstrates superior performance across all major embedding tasks -- including retrieval, classification and semantic textual similarity (STS) -- and excels in challenging multilingual scenarios, such as low-resource languages and cross-lingual setups. This state-of-the-art performance is driven by a novel data mix of 16.1 million query-document pairs, split between
7.7 million samples from public datasets and 8.4 million synthetically generated examples
from various open-weight LLMs. One of our key contributions is a detailed ablation study analyzing core design choices, including a comparison of contrastive loss implementations, an evaluation of synthetic data generation (SDG) strategies, and the impact of model merging. The llama-embed-nemotron-8b is an instruction-aware model, supporting user-defined instructions to enhance performance for specific use-cases. This combination of top-tier performance, broad applicability, and user-driven flexibility enables it to serve as a universal text embedding solution.

\end{abstract}

\section{Introduction} 

Dense text embedding models are a fundamental component of modern information retrieval. They are critical for a wide range of applications, including web search, question answering, semantic textual similarity, and recommendation engines. Their importance has been further amplified by the widespread adoption of Retrieval-Augmented Generation (RAG), which grounds Large Language Models (LLMs) in external context. Recent notable text embedding models include NV-Embed~\cite{nv-embed}, NV-Retriever~\cite{nv-retriever}, Qwen3-Embedding~\cite{qwen3-embed}, and Gemini~Embedding~\cite{gemini-embed}. These models achieve strong results on benchmarks like the Massive Text Embedding Benchmark (MTEB)~\cite{mteb,mtebnew}, which comprehensively evaluate models across a broad range of tasks.

In parallel, the field has seen a significant shift towards multi-modal~\cite{colpali,nemoretriever-colembed,jina-embeddings-v4} and omni-modal~\cite{omni-embed-nemotron} embedding models. This trend is driven by the need to handle real-world documents, such as PDFs or slides, which contain a mix of text, tables, and charts, as well as other rich modalities like audio and video. While multi-modal models address this trend, high-performance, text-only embedding models remain a relevant and efficient solution for a wide range of text-centric use cases. This includes a large volume of inherently text-native data, such as news articles, support tickets, and legal documents, as well as popular pipelines where documents like scanned PDFs or invoices are first converted to text via Optical Character Recognition (OCR).

The central challenge in this domain remains the development of a truly "universal" text model that performs robustly across diverse tasks, domains, and, critically, multiple languages. To address this challenge, we introduce llama-embed-nemotron-8b, a new universal text embedding model. Our model establishes a new state-of-the-art, achieving the 1st place rank on the comprehensive Multilingual Massive Text Embedding Benchmark (MMTEB) leaderboard~\cite{mtebnew} (as of October 21, 2025). See aggregated results in Table~\ref{tab:mmteb_aggregate}.

This paper details the architecture, training methodology, and data mixing strategies that enable llama-embed-nemotron-8b to effectively unify text representation across a wide spectrum of languages and text embedding tasks.

\begin{table*}
  \caption{Aggregated results for the MTEB(Multilingual, v2) split of the MTEB Leaderboard (as of October 21, 2025). Ranking on the Leaderboard is performed based on the Borda rank. Each task is treated as a preference voter, which gives votes to the models based on their relative performance on the task. The model with the highest number of votes across all tasks obtains the best rank.}
  \label{tab:mmteb_aggregate}
    \centering
    \small
  \begin{tabular}{lcccc}
    \toprule
    Model & Borda Rank & Borda Votes & \makecell{Mean (Task) \\ 131 tasks}  & \makecell{Mean (Task Type) \\ 9 task types} \\
    \midrule
llama-embed-nemotron-8b & \textbf{1.} & \textbf{39,573} & 69.46 & 61.09 \\
    \midrule
gemini-embedding-001 & 2. & 39,368 & 68.37 & 59.59 \\
Qwen3-Embedding-8B & 3. & 39,364 & \textbf{70.58} & \textbf{61.69} \\
Qwen3-Embedding-4B & 4. & 39,099 & 69.45 & 60.86 \\
Qwen3-Embedding-0.6B & 5. & 37,419 & 64.34 & 56.01 \\
gte-Qwen2-7B-instruct & 6. & 37,167 & 62.51 & 55.93 \\
Linq-Embed-Mistral & 7. & 37,149 & 61.47 & 54.14 \\
    \bottomrule
  \end{tabular}
\end{table*}

\section{Model}
\label{sec:model}

Our model, llama-embed-nemotron-8b, is a universal, instruction-tuned text embedding model designed to generate specialized embeddings for a wide range of tasks, including retrieval, classification, and STS. It has the ability to adapt its embedding outputs based on a task-specific instructional prefix.

We initialize the model using the weights and architecture of the Llama-3.1-8B model~\cite{meta_llama_3_1}. The base Llama-3.1 model is a decoder-only transformer that employs a causal attention mask, where each token can only attend to itself and previous tokens. We replace the causal attention mask in all transformer layers with a standard bi-directional attention (i.e., no masking). This allows every token in the input sequence to freely attend to all other tokens in the sequence, effectively converting the model into a bi-directional encoder. We unfreeze all the model weights and fine-tune Llama-3.1-8B end-to-end.

To produce a single, fixed-size embedding, the model processes the tokenized input sequence $S$ and produces a sequence of hidden states $H \in \mathbb{R}^{L \times d_{model}}$ from its final transformer layer, where $L$ is the sequence length and $d_{model}$ is the model's hidden dimension (4096). We then apply global average pooling over the sequence dimension of these final hidden states to obtain the final embedding vector $v$.

The model's specialization for diverse task types is guided by a textual instruction provided in the input. All inputs $T$ are formatted using a specific template:
\begin{center}
\begin{minipage}{0.65\linewidth}
\begin{verbatim}
Input = f"Instruct: {task_instruction}\nQuery: {T}"
\end{verbatim}
\end{minipage}
\end{center}
where $task\_instruction$ is a string that tells the model to produce an embedding suitable for the target task or use-case.

While the core encoder model is shared, its application architecture varies depending on the task family:
\begin{itemize}
    \item \textbf{For Retrieval Tasks:} We employ a bi-encoder architecture. The query and corpus (documents) are processed independently by the shared encoder. This projects them into a common embedding space. Query is using the appropriate instruction template, while no specific formatting is required for documents.  At inference time, relevance is computed using cosine similarity, enabling efficient and scalable search across large corpora.

    \item \textbf{For STS \& Classification Tasks:} The model functions as a uni-encoder. Each text is passed through the model with the appropriate instruction to generate its embedding. These embeddings are then directly used for the task, such as computing cosine similarity for STS or serving as features for a classifier.
\end{itemize}

This flexible, instruction-driven approach allows a single model to effectively handle the varied demands of all the MMTEB task types.

\section{Training}

\subsection{Training Objective}

We leverage contrastive learning to train the model, mapping inputs to a shared embedding space. The core objective is to maximize the similarity between related items and minimize it for unrelated ones. While the specific definition of the training triplet -- an anchor query ($q$), a positive document ($d^+$), and a set of negative documents ($D_N$) -- is adapted depending on the specific problem type, the training is governed by the InfoNCE contrastive loss~\cite{infonce}. The formal objective is:

\begin{equation}
\mathcal{L}(q, d^+, D_N) = -\log \frac{\exp(\text{sim}(q, d^+)/\tau)}{\sum_{d_i \in \{d^+\} \cup D_N} \exp(\text{sim}(q, d_i)/\tau)}
\label{eq:infonce}
\end{equation}

\noindent where $q$ is the embedding of an anchor, $d^+$ is the embedding of the positive item, and $D_N$ denotes the set of negative items. $sim(\cdot)$ represents the cosine similarity function, and $\tau$ is the temperature hyperparameter.

While Equation~\ref{eq:infonce} defines the core objective, the composition of the training triplet ($q, d^+, D_N$) is adapted for the different task types:

\begin{itemize}
    \item \textbf{For Retrieval Tasks:} The inputs directly map to a <query, positive document, negative documents> triplet. The query $q$ is formatted with the retrieval instruction (as described in Section~\ref{sec:model}), while documents ($d^+$ and $D_N$) do not require any prefix instructions. Some of the popular components which are frequently added to the $D_N$ set are in-batch negatives~\cite{gemini-embed} and same-tower negatives~\cite{qwen3-embed,gecko}. For our model training, we do not utilize any extra negatives in $D_N$, apart from the mined hard negatives. Our hard negative mining process is described in Section~\ref{sec:hnm}.

    \item \textbf{For Classification Tasks:} The input text serves as the anchor $q$. The positive $d^+$ is the text of the correct label name. The negatives $D_N$ are a set of random incorrect label names from the given classification task. The anchor is formatted with the specific classification instructions, while label names are processed without any modifications.

    \item \textbf{For STS Tasks:} These tasks are treated as symmetric. Given a positive pair of texts $(T_A, T_B)$, we create a training instance: $q=T_A, d^+=T_B$. In this case, $D_N$ contains hard negative examples which are mined from the dataset's corpus. All texts (query, positive, and negatives) are processed using the same instruction prefix: "Retrieve semantically similar text.".
\end{itemize}

\subsection{Training Stages}

We start training from Llama-3.1-8B foundation model weights~\cite{meta_llama_3_1}. Llama-3.1-8B model is already pre-trained on a corpus of about 15T multilingual tokens. This makes it a strong base model for training multi-lingual text embedding models. We train the model in two stages described below. Detailed hyperparameters for each stage are provided in Appendix~\ref{sec:hparams}.

\paragraph{Stage 1: Retrieval Pretraining.}
The goal of the first stage is to adapt Llama-3.1-8B LLM to both bi-directional attention and embedding model setup. In this stage we use only retrieval data where queries and documents are based on the Web corpus. We use only a single hard-negative for each <query, document> pair which is mined from the same Web corpus. This stage constitutes about 70\% of the overall data mix.

\paragraph{Stage 2: Fine-Tuning.}
In the second stage, we fine-tune the model using high-quality datasets for various problem types: retrieval, classification, STS, and bitext mining. The goal of this stage is to train a well-rounded model that works across different tasks. This stage constitutes the other 30\% of the overall data mix.

\subsection{Model Merging}
\label{sec:model-merging}
We train multiple models using the two-stage approach above, and then apply model merging across resulting checkpoints. Model merging involves combining the parameters of multiple models, and is a well-established technique \cite{model-merge-2018, model-merge-2022}. Recently, this approach has been popularized for improving the robustness and generalization of embedding models, such as Qwen3-Embedding~\cite{qwen3-embed}, Gemini~Embedding~\cite{gemini-embed}, and EmbeddingGemma~\cite{embedding-gemma}.

The core idea of this method is to average the parameters obtained from the individual model runs. There are different strategies for selecting the individual checkpoints. These include averaging checkpoints from different steps within the same training run~\cite{model-merge-2018}, or combining models from multiple, distinct training runs that may use different hyperparameters or intentional data variations~\cite{model-merge-2022}.

Our final model is an average of six diverse individual checkpoints. We achieved this diversity by varying the data mixes and model hyperparameters across training runs. This final model produces the best evaluation results compared to individual checkpoints, with no inference time increase. We provide more details about the results in the ablation studies (Section \ref{sec:abl-model-merging}).

\section{Datasets}

This section provides details about our data curation process. We plan to release our data mix, it will be available as a part of our HuggingFace collection\footnote{Nemotron RAG collection: https://huggingface.co/collections/nvidia/nemotron-rag}. The whole data mix contains a total of 16 million <query, document> pairs. Its overview is presented in Table~\ref{tab:data_aggregate}.

\begin{table*}
  \caption{Overview of the training data mix, detailing the number of <query, document> pairs for pretraining and fine-tuning, segmented by non-synthetic and synthetic data sources.}
  \label{tab:data_aggregate}
    \centering
    \small
  \begin{tabular}{l|cc}
    \toprule
    Training Stage & Non-Synthetic Data & Synthetic Data \\
    \midrule
Pretraining & 5.0M & 6.8M  \\
Fine-tuning & 2.7M & 1.6M  \\
    \midrule
Total & 7.7M & 8.4M \\
    \bottomrule
  \end{tabular}
\end{table*}

\subsection{Pretraining Data Mix}

For pretraining data, we relied on the NVIDIA's Nemotron-CC-v2 dataset~\cite{nemotron-cc-v2}. We employed two strategies to create our final pretraining set, which consists of approximately 11.8M <query,~document> pairs.

\paragraph{Utilize existing questions from Nemotron-CC-v2.}
In this strategy, we utilized the Diverse-QA split of the Nemotron-CC-v2 dataset. We extracted the existing questions and their corresponding positive documents from this split. We then mined hard negatives from a pool of 1M document chunks sampled from the same Diverse-QA corpus. This approach yielded 5.0M training pairs.

\paragraph{Generate new questions for Nemotron-CC-v2 corpus.}
For our second strategy, we generated new synthetic queries. We took the existing documents from the Diverse-QA split and generated our own synthetic questions for them, creating a new set of <query, positive document> pairs. Subsequently, we mined hard negatives for these new pairs in the same manner as the first strategy. This approach contributed the remaining 6.8M training pairs.

\subsection{Fine-tuning Data Mix}

For the fine-tuning data mix, we started with a mix introduced in NV-Embed~\cite{nv-embed}. The MTEB Leaderboard~\cite{mtebnew} reports a zero-shot percentage for each model, which indicates whether any of the benchmark's train, validation, or test splits were used during a model's training phase. This metric is designed to ensure that the MTEB evaluation datasets remain out-of-domain for the models being tested. Therefore, to preserve the integrity of our zero-shot evaluation, we removed the majority of data originating from both MTEB(Multilingual, v2) and MTEB(eng, v2) splits of the MTEB.

As reported in Table~\ref{tab:data_aggregate}, our fine-tuning mix consists of two parts: non-synthetic and synthetic data. For non-synthetic part, we utilized well-known public datasets, like MIRACL~\cite{miracl}, HotpotQA~\cite{hotpotqa}, MS~MARCO\cite{ms-marco}, Natural~Questions~\cite{nq}, SQuAD~\cite{squad}, and more. The next section describes the synthetic part. The full list of fine-tuning datasets, together with a number of samples is presented in Appendix~\ref{sec:finetuning-data-mix}.

\subsection{Synthetic Data Generation}

To enhance the diversity of our datasets, we employed a comprehensive Synthetic Data Generation (SDG) strategy, with a specific focus on multi-lingual and cross-lingual data. We applied SDG for creating datasets across primary task types: retrieval, classification, STS, and bitext mining.

Our methodology relied on two main strategies, inspired by the recent state-of-the-art embedding models. The first strategy, similar to~\cite{e5-mistral} approach, involved the end-to-end generation of complete <query, positive, negatives> text triplets from scratch. The second strategy, inspired by~\cite{qwen3-embed} and~\cite{gemini-embed}, leveraged a seed corpus. We first sampled a positive document from the corpus, generated a corresponding query, and subsequently mined hard-negatives from the same corpus.

Additionally, we expanded our multi-lingual data by translating several existing high-quality datasets into various target languages. For all SDG and translation tasks, we utilized a diverse mix of powerful, open-weights LLMs. This list includes: gpt-oss-20b and gpt-oss-120b~\cite{openai2025gptoss120bgptoss20bmodel}, Mixtral-8x22B-Instruct-v0.1~\cite{jiang2024mixtralexperts}, Llama-3.3-70B-Instruct~\cite{meta_llama_3_1}, Llama-4-Scout-17B-16E-Instruct and Llama-4-Maverick-17B-128E-Instruct~\cite{meta_ai_llama4_2025}. We further explore the quality of synthetic datasets produced by different LLMs in our ablation studies (Section \ref{sec:ablation-llm}).

\subsection{Hard Negative Mining}
\label{sec:hnm}

To improve the effectiveness of contrastive learning, we incorporated the \textit{top-k with percentage to positive threshold} strategy from NV-Retriever~\cite{nv-retriever} for hard negative mining. For this process, we set the threshold at 0.95. This means that for a given query, we selected the top $K$ most relevant negative samples whose similarity to the query is less than 95\% of the query–positive similarity score. This approach encourages the model to learn from challenging negatives while simultaneously filtering out potential false negatives that have high similarity scores. We sourced hard negatives using a combination of two embedding models: e5-mistral-7b-instruct~\cite{e5-mistral} and Qwen3-Embedding-8B~\cite{qwen3-embed}.

\section{Results}

We evaluate our model on the Multilingual split of the MTEB benchmark~\cite{mteb}, which was introduced in Massive Multilingual Text Embedding Benchmark (MMTEB)~\cite{mtebnew}. This is the most extensive benchmark for multilingual and cross-lingual text embedding models, comprising 131 diverse tasks across 9 task types and 250+ languages (both high- and low-resource). The task types include Bitext Mining, Classification, Clustering, Instruction Reranking, Multilabel Classification, Pair Classification, Reranking, Retrieval, and STS.

Our model is instruction-aware, supporting custom instructions to optimize performance for specific use cases. For the MMTEB evaluations, we firstly took task-specific instructions directly from the MMTEB evaluation datasets, and adapted instructions from the Qwen3-Embedding evaluation repository~\footnote{Qwen3-Embedding GitHub repository: https://github.com/QwenLM/Qwen3-Embedding} for tasks without default instructions.

We present a detailed comparison of our model against the other top-10 models on the MMTEB Leaderboard (as of October 21, 2025) in Table~\ref{tab:mmteb_detail}. Other models in the comparison include gemini-embedding-001~\cite{gemini-embed}, Qwen3-Embedding family of models~\cite{qwen3-embed}, gte-Qwen2-7B-instruct~\cite{li2023towards}, Linq-Embed-Mistral~\cite{LinqAIResearch2024}, multilingual-e5-large-instruct~\cite{wang2024multilingual}, embeddinggemma-300m~\cite{embedding-gemma} and SFR-Embedding-Mistral~\cite{SFRAIResearch2024}. 

Our model achieves state-of-the-art performance, securing the Rank 1 position with 39,573 Borda votes. This represents a significant lead of over 200 votes compared to the 2nd place (gemini-embedding-001) and 3rd place (Qwen3-Embedding-8B) models.

Ranking on the official MMTEB Leaderboard is determined by the Borda count method~\cite{mtebnew}. Each task is treated as a preference voter, which gives votes to the models based on their relative performance on the task. The best model obtains the highest number of votes. The model with the highest number of votes across all tasks obtains the highest rank. The Borda count method has been shown to be more robust for comparing NLP systems~\cite{borda}. While Qwen3-Embedding-8B, achieves a higher "Mean (Task)" score (70.58 vs. our 69.46), this mean metric can be sensitive to outlier performance on a small subset of benchmarks. High scores on a few tasks can inflate the overall average without necessarily indicating consistent generalization. In contrast, the Borda rank is designed to reward broad and consistent generalization across the entire spectrum of 131 tasks, rather than strong performance in a limited number of areas.

\begin{table*}
\centering
\tiny
  \caption{Evaluation results of top leaderboard models on MTEB(Multilingual, v2) Leaderboard (as of October 21, 2025). Ranking on the official Leaderboard is determined by the Borda rank. Mean (Task) column is the average of scores across 131 individual tasks, while Mean (Type) is the average across 9 problem types.}
\setlength{\tabcolsep}{5pt}
\label{tab:mmteb_detail}
\begin{tabular}{lcccc|ccccccccc}
\toprule
 Model & \makecell{Borda \\ Rank} & \makecell{Borda \\ Votes} & \makecell{Mean \\(Task)} & \makecell{Mean \\ (Type)} & \makecell{Bitext \\ Mining} & \makecell{Class.} & \makecell{Clust.} & \makecell{Instr. \\ Rerank.} & \makecell{Multi. \\ Class.} & \makecell{Pair \\ Class.} & \makecell{Rerank.} & Retrieval & STS \\
\midrule
 \textbf{llama-embed-nemotron-8b} & \textbf{1.} & \textbf{39,573} & 69.46 & 61.09 & \textbf{81.72} & 73.21 & 54.35 & 10.82 & \textbf{29.86} & 83.97 & \textbf{67.78} & 68.69 & 79.41 \\
 \midrule
 gemini-embedding-001 & 2. & 39,368 & 68.37 & 59.59 & 79.28 & 71.82 & 54.59 & 5.18 & 29.16 & 83.63 & 65.58 & 67.71 & 79.40 \\
 Qwen3-Embedding-8B & 3. & 39,364 & \textbf{70.58} & \textbf{61.69} & 80.89 & \textbf{74.00} & \textbf{57.65} & 10.06 & 28.66 & \textbf{86.40} & 65.63 & \textbf{70.88} & \textbf{81.08} \\
 Qwen3-Embedding-4B & 4. & 39,100 & 69.45 & 60.86 & 79.36 & 72.33 & 57.15 & \textbf{11.56} & 26.77 & 85.05 & 65.08 & 69.60 & 80.86 \\
 Qwen3-Embedding-0.6B & 5. & 37,419 & 64.34 & 56.01 & 72.23 & 66.83 & 52.33 & 5.09 & 24.59 & 80.83 & 61.41 & 64.65 & 76.17 \\
 gte-Qwen2-7B-instruct & 6. & 37,167 & 62.51 & 55.93 & 73.92 & 61.55 & 52.77 & 4.94 & 25.48 & 85.13 & 65.55 & 60.08 & 73.98 \\
 Linq-Embed-Mistral & 7. & 37,149 & 61.47 & 54.14 & 70.34 & 62.24 & 50.60 & 0.94 & 24.77 & 80.43 & 64.37 & 58.69 & 74.86 \\
 multilingual-e5-large-instruct & 8. & 36,921 & 63.22 & 55.08 & 80.13 & 64.94 & 50.75 & -0.40 & 22.91 & 80.86 & 62.61 & 57.12 & 76.81 \\
 embeddinggemma-300m & 9. & 36,728 & 61.15 & 54.31 & 64.40 & 60.90 & 51.17 & 5.61 & 24.82 & 81.40 & 63.25 & 62.49 & 74.73 \\
 SFR-Embedding-Mistral & 10. & 36,579 & 60.90 & 53.92 & 70.00 & 60.02 & 51.84 & 0.16 & 24.55 & 80.29 & 64.19 & 59.44 & 74.79 \\
\bottomrule
\end{tabular}
\end{table*}

\section{Ablation Study}
\label{sec:ablation}

In this section, we ablate design choices that helped the development of the llama-embed-nemotron-8b model. Due to the computational cost of ablating at the 8B scale, all studies presented below (unless mentioned otherwise) were conducted on a 1B model fine-tuned from Llama-3.2-1B~\cite{meta_llama_3_1} on our fine-tuning data mix. These smaller-scale experiments allowed us to validate our training decisions before scaling up the experiments.

\subsection{Contrastive Loss Formulations}

In this analysis, we compare our InfoNCE loss implementation to other prominent formulations. These approaches primarily differ in the composition of negative samples used in the loss denominator.

\begin{itemize}
    \item Gecko Model~\cite{gecko} implementation contrasts a <query, positive passage> pair against a comprehensive set of negatives: (1) a single hard negative, (2) other positive passages from different queries in the batch, and (3) other queries in the batch. This third category is termed "same-tower negatives"~\cite{moiseev2023samtoneimprovingcontrastiveloss}, which are noted as being beneficial for symmetric text embedding tasks (e.g., semantic similarity).
    
    \item Qwen3-Embedding family of models~\cite{qwen3-embed} uses same-tower negatives not only for the queries, but also for in-batch positive and negative documents.
    
    \item Gemini Embedding~\cite{gemini-embed} explicitly omits the same-tower negatives from the loss to avoid potential false negatives. The loss denominator is thus limited to only a single hard negative and other positive passages in the batch.
    
    \item \textbf{Ours:} In contrast, our approach simplifies the loss denominator to include only hard negative documents (HNs) (one in pretraining, four in fine-tuning). This formulation omits all the in-batch negatives and same-tower negatives.
\end{itemize}

To have a fair comparison, we have fixed all the hyperaparameters, and only tuned a learning rate for each loss separately. As shown in Table~\ref{tab:loss}, all approaches achieve similar performance. This suggests that the inclusion of in-batch negatives or same-tower negatives provides minimal-to-no significant benefit over our simpler approach. Our implementation, which relies only on hard negatives, achieves the highest number of Borda votes (38,225) and wins the most individual task types.

\begin{table*}
\centering
\tiny
  \caption{Comparison of different InfoNCE loss implementations on the MMTEB Leaderboard.}
\setlength{\tabcolsep}{6pt}
\label{tab:loss}
\begin{tabular}{lccc|ccccccccc}
\toprule
 Loss & \makecell{Borda \\ Votes} & \makecell{Mean \\(Task)} & \makecell{Mean \\ (Type)} & \makecell{Bitext \\ Mining} & \makecell{Class.} & \makecell{Clust.} & \makecell{Instr. \\ Rerank.} & \makecell{Multi. \\ Class.} & \makecell{Pair \\ Class.} & \makecell{Rerank.} & Retrieval & STS \\
\midrule
 Gecko & 37,903 & 63.45	 & 55.86	 & 72.58 & 	65.12	 & 52.29 & 	5.28	 & 25.46 & 	80.68	 & 64.21	 & \textbf{61.20}	 & 75.87 \\

 Qwen3-Embedding & 36,835 & 62.14	 & 55.49 & 	\textbf{73.50}	 & 60.60 & 	\textbf{54.82}	 & 4.97	 & 25.22	 & 80.63 & 	64.11	 & 60.12 & 	75.41 \\
 Gemini Embedding & 38,135 & 63.83 & 	55.90 & 	73.13 & 	66.28 & 	53.04 & 	4.76 & 	\textbf{24.70} & 	80.74	 & 64.26	 & 60.28	 & 75.96 \\
 Ours (HNs Only) & \textbf{38,225} & \textbf{64.03} & 	\textbf{56.04}	 & 72.94	 & \textbf{66.99}	 & 52.27	 & \textbf{5.69}	 & 24.60	& \textbf{80.77}	 & \textbf{64.44}	 & 60.66	 & \textbf{75.99} \\
\bottomrule
\end{tabular}
\end{table*}

\subsection{Choice of LLM for Synthetic Data Generation}
\label{sec:ablation-llm}

For training llama-embed-nemotron-8b we relied on multiple open-weights LLMs to generate synthetic data for various problem types: retrieval, classification, STS, and bitext mining. This analysis focuses on evaluating LLMs for the task of generating classification datasets. Following~\cite{nv-embed, e5-mistral}, we created synthetic examples in 2 steps:

\begin{itemize}
    \item Step 1. Prompt LLM to generate a list of potential classification tasks. These tasks are also used as instructions in our instruction-aware training.
    \item Step 2. Given the classification task, prompt LLM to generate (a) a text sample, (b) the correct label, and (c) a list of plausible but incorrect labels (misleading labels). We use the correct label name as a positive and the misleading label names as negatives.
\end{itemize}

Our baseline is a model that was trained without any synthetic classification datasets. We compare it against generating 100k synthetic samples using each of the LLMs, and training separate embedding models with extra 100k examples in the data mix. For each LLM we follow Step~1 and Step~2 described above, and generate data only in English language. LLMs being evaluated include gpt-oss-20b and gpt-oss-120b~\cite{openai2025gptoss120bgptoss20bmodel}, Mixtral-8x22B-Instruct-v0.1~\cite{jiang2024mixtralexperts}, Llama-3.3-70B-Instruct~\cite{meta_llama_3_1}, Llama-4-Scout-17B-16E-Instruct and Llama-4-Maverick-17B-128E-Instruct~\cite{meta_ai_llama4_2025}.

By comparing performance on individual MMTEB evaluation datasets, we observed that different LLMs for SDG excel in different domains/languages. Therefore, we also compare the performance of each individual LLM with another mix of 100k synthetic samples, which mixes examples from all the LLMs with equal weights (i.e., $\approx$16.7k samples from each of the 6 models). See our full results in Table~\ref{tab:sdg-llm}.

We compare results on multiple task types including classification, multilabel classification and clustering, which is also closely related to the classification. Results suggest that the largest LLM is not necessarily the best model for SDG, as gpt-oss-20b performs very well, while being the smallest model. But the best results are achieved by using a data mix compiled from all the LLMs. These findings suggest that diversity of synthetic data is more important than single-model quality. One of the reasons for such behavior might be that Mix approach has a more diverse tasks list from the Step~1 compared to individual LLMs.

This mixing approach for SDG is used in both our pretraining and fine-tuning datasets, and across all the problem types. We also extend this principle by using cross-model SDG, where Step~1 classification tasks are generated with one model, while Step~2 actual samples are generated with another model.

To conclude this analysis, we compare a baseline without synthetic classification data to the best SDG approach. We can see how only 100k synthetic classification examples show an improvement of +464 Borda votes (37,812 vs 37,348) and +0.94 in Mean points (62.89 vs 61.95). This demonstrates the efficacy of synthetic data. The next ablation explores how good synthetic data is, compared to in-domain classification datasets.

\begin{table*}
  \caption{Comparison of different LLMs for generating synthetic classification datasets.}
\label{tab:sdg-llm}
    \centering
    \setlength{\tabcolsep}{6pt}
    \scriptsize
  \begin{tabular}{lc|ccccc}
    \toprule
    Model for SDG & \makecell{Number of \\ Parameters} & \makecell{Borda \\ Votes} & \makecell{Mean \\ (Task)} & Class.  & Clust. & \makecell{Multi. \\ Class.} \\
    \midrule
No synthetic data & - & 37,348 & 61.95 & 62.16 & 49.94 & 22.40 \\
    \midrule
gpt-oss-20b & 21B (3.6B active) & 37,732 & 62.54 & 	\underline{63.71} & 	50.45 & 	\underline{23.21} \\
gpt-oss-120b & 117B (5.1B active) & 37,594 & 62.38	 & 63.12	 & 50.77 & 	22.47 \\
Mixtral-8x22B-Instruct-v0.1 & 141B (39B active) & \underline{37,797} & \underline{62.64} & 	63.67 & 	\underline{50.89} & 	22.81 \\
Llama-3.3-70B-Instruct & 70B & 37,623 & 62.49 & 	63.15 & 	50.80 & 	22.85 \\
Llama-4-Scout-17B-16E-Instruct & 109B (17B active) & 37,595 & 62.21 & 	63.02 & 	50.57 & 	22.70 \\
Llama-4-Maverick-17B-128E-Instruct & 400B (17B active) & 37,643 & 62.36 & 	63.20 & 	50.51 & 	22.41 \\
Mix from all models & - & \textbf{37,812} & \textbf{62.89} & 	\textbf{64.39} & 	\textbf{50.95} & 	\textbf{23.37} \\
    \bottomrule
  \end{tabular}
\end{table*}

\subsection{Impact of Synthetic vs. In-Domain Data}

The MTEB leaderboard~\cite{mtebnew} tracks whether models use in-domain data in their data mixes (e.g., the training split of an evaluation dataset). This "zero-shot" percentage is tracked because in-domain data can significantly inflate performance on a specific evaluation dataset, making comparisons difficult.

This ablation study quantifies the gap between our synthetic data and in-domain data. The goal is to measure the effectiveness of our synthetic mix in closing the performance gap to a model trained on in-domain data. For this analysis, we randomly selected five classification evaluation datasets from MTEB(Multilingual, v2), namely AmazonCounterfactualClassification~\cite{oneill-etal-2021-wish}, CzechProductReviewSentimentClassification~\cite{habernal-etal-2013-sentiment}, GreekLegalCodeClassification~\cite{papaloukas-etal-2021-glc}, EstonianValenceClassification~\cite{Pajupuu2023}, and TweetTopicSingleClassification~\cite{dimosthenis-etal-2022-twitter}.

The baseline model is a model trained without any synthetic classification datasets. It is compared to two other models. Crucially, the model trained on in-domain data described below was prepared solely for this ablation study to serve as a comparative benchmark. This in-domain data was not used in our final llama-embed-nemotron-8b model submitted to the MTEB leaderboard.

\begin{itemize}
    \item First model is trained on about 1M synthetic classification samples generated with the approach described in Section~\ref{sec:ablation-llm}.
    \item Second model is trained on train splits of all the five datasets being evaluated: AmazonCounterfactualClassification (17.7k observations), CzechProductReviewSentimentClassification (24.0k), GreekLegalCodeClassification (28.5k), EstonianValenceClassification (3.3k) and TweetTopicSingleClassification (1.5k).
\end{itemize}

Results are presented in Table~\ref{tab:in-domain}. The model trained on our synthetic classification data consistently outperforms baseline model across all five tasks. However, model trained on in-domain data achieves the highest scores, substantially outperforming the synthetic data mix. Notably, even a very small amount of in-domain data provides a powerful signal, with 1.5k train samples from TweetTopicSingleClassification dataset surpassing about 1M synthetic samples.

This shows that while our synthetic data allows to improve general performance on classification benchmarks, it is not a complete substitute for acquiring even small amounts of high-quality, in-domain data.

\begin{table*}
  \caption{Comparison of synthetic datasets against in-domain data.}
\label{tab:in-domain}
    \centering
    \setlength{\tabcolsep}{5pt}
    \small
  \begin{tabular}{l|ccccc}
    \toprule
    Data & Amazon & Czech & Greek  & Estonian & TweetTopic \\
    \midrule
Baseline (no synthetic data) & 73.06 & 64.68 & 38.21 & 38.80 & 72.85 \\
    \midrule
+1M synthetic samples & 82.64 & 66.58 & 	42.60 & 	49.36 & 	78.00 \\
+75k in-domain samples & \textbf{90.51} & \textbf{71.73}	 & \textbf{60.51}	 & \textbf{57.20} & 	\textbf{80.12} \\
    \bottomrule
  \end{tabular}
\end{table*}

\subsection{Model Merging}
\label{sec:abl-model-merging}

In Section~\ref{sec:model-merging}, we discussed a model merging technique that enhances model's generalizability at no additional inference costs. This ablation quantifies the impact of this technique by comparing the merged model against its individual component checkpoints. Our final llama-embed-nemotron-8b model is an average of six diverse individual models, weighted equally. This diversity stems from varying data mixes and hyperparameter sets used during training.

In Table~\ref{tab:model-merging}, we present the MTEB(Multilingual, v2) results for the individual checkpoints and the final merged model. Notably, our best individual model ("Model 6") would already achieve SOTA performance on the MMTEB Leaderboard, securing 39,454 Borda votes (as of October 21, 2025). However, model merging yields a significantly stronger model, improving the score by +119 Borda votes (39,573 vs 39,454) and mean by +0.84 (69.46 vs 68.62) over "Model 6".

A key observation is that the individual models specialize in different task types. For instance, "Model 4" specializes in clustering and reranking; "Model 5" -- in pair classification and retrieval; "Model 6" -- in classification and STS. Merging all six checkpoints together creates a robust cumulative ensemble that aggregates these complementary strengths. This results in the strongest overall model, which achieves the top score in almost all problem types (apart from clustering).

\begin{table*}
\centering
\tiny
  \caption{Evaluation results of individual checkpoints and the final llama-embed-nemotron-8b model on the MTEB(Multilingual, v2) Leaderboard. llama-embed-nemotron-8b is an average of six individual models listed in the table.}
\setlength{\tabcolsep}{6pt}
\label{tab:model-merging}
\begin{tabular}{lccc|cccccccccc}
\toprule
 Model & \makecell{Borda \\ Votes} & \makecell{Mean \\(Task)} & \makecell{Mean \\ (Type)} & \makecell{Bitext \\ Mining} & \makecell{Class.} & \makecell{Clust.} & \makecell{Instr. \\ Rerank.} & \makecell{Multi. \\ Class.} & \makecell{Pair \\ Class.} & \makecell{Rerank.} & Retrieval & STS \\
\midrule
 Individual model 1 & 39,167 & 67.27 & 59.36 & 78.45 & 70.17 & 54.11 & 10.28 & 27.86 & 83.25 & 66.84 & 64.99 & 78.27 \\
  Individual model 2 & 39,265 & 67.99 & 59.78 & 78.60 & 71.83 & 53.79 & 10.42 & 28.59 & 83.86 & 66.85 & 65.77 & 78.33 \\
  Individual model 3 & 39,336 & 68.00 & 59.71 & 79.33 & 71.78 & 54.27 & 9.14 & 28.64 & 83.76 & 66.83 & 64.86 & 78.82 \\
  Individual model 4 & 39,401 & 68.36 & 60.18 & 79.52 & 71.89 & \textbf{54.91} & 10.32 & 29.20 & 83.82 & \underline{67.04} & 66.15 & 78.76 \\
  Individual model 5 & 39,435 & 68.38 & 60.05 & 79.37 & 71.87 & 54.12 & 9.48 & 28.72 & \underline{83.91} & 66.85 & \underline{67.33} & 78.83 \\
  Individual model 6 & \underline{39,454} & \underline{68.62} & \underline{60.37} & \underline{79.56} & \underline{72.37} & 54.34 & \underline{10.80} & \underline{29.71} & 83.60 & 66.91 & 66.99 & \underline{79.08} \\
 \midrule
 llama-embed-nemotron-8b & \textbf{39,573} & \textbf{69.46} & \textbf{61.09} & \textbf{81.72} & \textbf{73.21} & \underline{54.35} & \textbf{10.82} & \textbf{29.86} & \textbf{83.97} & \textbf{67.78} & \textbf{68.69} & \textbf{79.41} \\

\bottomrule
\end{tabular}
\end{table*}

\section{Conclusion}

In this work, we introduced llama-embed-nemotron-8b, a new open-weights universal text embedding model. Our model achieves state-of-the-art performance, securing the \#1 position on the MMTEB leaderboard as of October 21, 2025. It demonstrates superior generalization according to the Borda count method, outperforming other top models across a diverse set of 131 tasks and 250+ languages.

This result is driven by a combination of a strong Llama-3.1-8B foundational model converted to a bi-directional encoder, a novel 16M-pair training data mix, and a robust training methodology. Our ablation studies revealed that using a mix of diverse open-weights LLMs for synthetic data generation yields more robust results than using any single LLM, emphasizing the importance of data diversity.

By releasing the weights of the llama-embed-nemotron-8b model, we provide a powerful, instruction-aware tool for a wide range of applications, including retrieval, classification, and STS. We also plan to release our curated data mix to facilitate future research and development in robust, multilingual text embeddings.
% \clearpage
\bibliographystyle{unsrt}
\bibliography{bibliography}

\clearpage

\appendix

\section{Implementation Details}
\label{sec:hparams}

This section details the hyperparameters and implementation specifics for our llama-embed-nemotron-8b model. Pretraining was conducted for 25.0 hours, and fine-tuning for 21.5 hours. Both stages utilized a cluster of 64 NVIDIA A100 80GB GPUs. The hyperparameters for both training stages are summarized in Table~\ref{tab:hparams}.

\begin{table}[H]
  \caption{Main hyperparameters for llama-embed-nemotron-8b training.}
  \label{tab:hparams}
    \centering
    \small
  \begin{tabular}{l|cc}
    \toprule
    Hyperparameter & Pretraining & Fine-tuning \\
    \midrule
Peak learning rate & 1e-5 & 2e-6  \\
Batch size & 2,048 & 128  \\
Number of steps & 5,773 & 33,668  \\
Scheduler & Linear decay  & Linear decay  \\
Warm-up steps & 100  & 100  \\
Optimizer & AdamW  & AdamW  \\
Weight decay & 0.01  & 0.01  \\
Number of hard negatives & 1  & 4  \\
Temperature & 0.02  & 0.02  \\
Query max length & 512  & 512  \\
Document max length & 512  & 512  \\
    \bottomrule
  \end{tabular}
\end{table}

\section{Fine-Tuning Data Mix for llama-embed-nemotron-8b}
\label{sec:finetuning-data-mix}

This section details the high-quality, curated data mix used in the fine-tuning stage of the llama-embed-nemotron-8b training. The complete dataset consists of 4.3 million samples, sourced from a diverse range of corpora, including multi-lingual and cross-lingual data.

The mix is composed of approximately 2.7 million non-synthetic samples from public sources and 1.6 million synthetic samples generated to target specific model capabilities. A detailed breakdown of the component datasets and their respective sample sizes is provided in Table~\ref{tab:data-mix}.

\begin{table}[H]
  \caption{Component datasets and sample counts for the llama-embed-nemotron-8b fine-tuning data mix.}
    \setlength{\tabcolsep}{15pt}
  \label{tab:data-mix}
    \centering
    \small
  \begin{tabular}{lr}
    \toprule
    Dataset & Number of samples \\
    \midrule
    AmazonReviews~\cite{mcauley-2013-hiddenfactors} & 60,000 \\
    BioASQ~\cite{tsatsaronis2015overview} & 2,495 \\
    DBLP-Citation-network V17~\cite{ArnetMiner} & 25,000 \\
    EmotionClassification~\cite{saravia-etal-2018-carer} & 13,046 \\
    FEVER~\cite{thorne2018fever} & 140,085 \\
    GooAQ~\cite{khashabi2021gooaqopenquestionanswering} & 100,000 \\
    HotpotQA~\cite{hotpotqa} & 170,000 \\
    HoVer~\cite{jiang2020hover} & 29,721 \\
    InF-IR~\cite{zhuang2025betterinstructionfollowingretrieval} & 77,518 \\
    MAmmoTH2 stackexchange~\cite{yue2024mammoth2} & 317,180 \\
    MIRACL~\cite{miracl} & 79,648 \\
    MLDR~\cite{bge-m3} & 9,500 \\
    Mr.TyDi~\cite{zhang2021mr} & 12,610 \\
    MS~MARCO\cite{ms-marco} & 500,000 \\
    MultiNLI~\cite{williams2018broadcoveragechallengecorpussentence} & 75,505 \\
    Natural~Questions~\cite{nq} & 100,231 \\
    NFCorpus~\cite{boteva2016} & 3,685 \\
    PAQ~\cite{lewis2021paq} & 500,000 \\
    Quora question pairs~\cite{quora-question-pairs} & 101,762 \\
    RedditClustering~\cite{geigle:2021:arxiv} & 90,000 \\
    SciFact~\cite{wadden2022scifact} & 919 \\
    SQuAD~\cite{squad} & 87,599 \\
    Stack Exchange~\cite{stack-exchage} & 80,001 \\
    SuperGLUE Textual Entailment~\cite{wang2020supergluestickierbenchmarkgeneralpurpose} & 3,094 \\
    Synthetic bitext mining data~\cite{Pang+Lee:05a} & 169,534 \\
    Synthetic classification data & 1,044,212 \\
    Synthetic retrieval data & 182,814 \\
    Synthetic STS data & 239,997 \\
    Toxic Comment Classification~\cite{jigsaw-toxic-comment-classification-challenge} & 16,800 \\
    TriviaQA~\cite{2017arXivtriviaqa} & 73,346 \\
    TuPy~\cite{silly-machine_2023} & 3,200 \\
    \midrule
    \textbf{Total} & \textbf{4,309,502} \\
    \bottomrule
  \end{tabular}
\end{table}
\end{document}